\documentclass[10pt, twocolumn, a4paper]{article}

\usepackage{cmap}
\usepackage[utf8]{inputenc}
\usepackage[T1]{fontenc}
\usepackage{lmodern}

\usepackage{geometry}
\usepackage{amsmath}
\usepackage{amssymb}
\usepackage{titlesec}
\usepackage{abstract}
\usepackage{graphicx}
\usepackage{hyperref}
\usepackage{caption}
\usepackage{subcaption}
\usepackage{array}

\begin{document}
	
	\twocolumn[{
	\begin{center}
		\includegraphics[height=1.5cm]{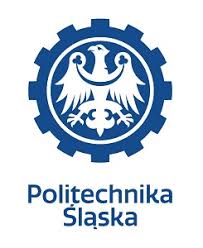} \hfill \includegraphics[height=1.5cm]{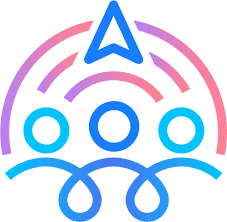} \\
		\vspace{1.5em}
		
		{\Large \textbf{MRI-based Deep Radiomic Phenotyping of Neuromuscular Disorders: A Topology-driven Characterization} \par}
		\vspace{1.5em}
		
		{\large 
			Martyna Żur$^1$, Łukasz Piórecki$^1$, Marek Socha$^1$, Jose Verdu Diaz$^2$, \\
			Ana Topf$^2$, Michael Obach$^4$, Marco Savarese$^5$, Jocelyn Laporte$^6$, \\
			Benedikt Schoeser$^7$, Filippo Santorelli$^8$, Jordi Diaz-Manera$^2$, \\
			Volker Straub$^2$, Rossella Tupler$^3$, Joanna Polańska$^1$ \par}
		\vspace{1em}
		
		{\small 
			$^1$Department of Data Science and Engineering, Silesian University of Technology, Gliwice, Poland \\
			$^2$John Walton Muscular Dystrophy Research Centre, Newcastle University, Newcastle upon Tyne, UK \\
			$^3$Department of Life Sciences, University of Modena and Reggio Emilia, Modena, Italy \\
			$^4$Fundación Tecnalia Research \& Innovation, Donostia-San Sebastián, Spain \\
			$^5$University of Helsinki, Helsinki, Finland \\
			$^6$University of Strasbourg, Strasbourg, France \\
			$^7$Deutsche Gesellschaft für Neurologie, European Academy of Neurology, Ludwig-Maximilians-Universität München Medizinische Fakultät, Munich, Germany \\
			$^8$Fondazione Stella Maris, Pisa, Italy \par}
		\vspace{1.5em}
	\end{center}
		
		\begin{onecolabstract}
			\textbf{Abstract:} Quantitative assessment of muscle MRI is crucial for monitoring neuromuscular disorders (NMD). This study introduces an automated radiomic phenotyping framework based on original features engineered across five main architectural domains: quantitative morphometry, spatial distribution, geometric shape, interactions between progressive fat replacement stages, and graph-based topology. Utilizing 1184 MRI scans from the CoMPaSS-NMD project, we map the complex 3D architecture of heterogeneous intramuscular lipodegeneration into objective, morphologically interpretable biomarkers. We introduce a graph-based skeletonization of fat infiltrates to quantify muscle architectural changes, establishing a multi-dimensional extension of traditional, spatially-agnostic volume metrics by mapping topological networks across the entire 3D muscle volume. Statistical screening via non-parametric Kruskal-Wallis analysis confirmed the discriminative power of these novel descriptors across the genetic hierarchy. Notably, topological network metrics (e.g., SF1\_Skel\_Nodes, $\epsilon^2 = 0.2656$) and interface dynamics metrics (e.g., SF2\_To\_SF1\_Dist\_Min, $\epsilon^2 = 0.2092$) demonstrated substantial effect sizes, providing deeper structural insights than classical volumetric assessments. Post-hoc pairwise evaluations and UMAP projections further indicated the capability of these topological and 3D geometric invariants to capture disease-specific macroscopic infiltration patterns. These results demonstrate that global architectural features represent a highly promising class of biomarkers for differential diagnosis, offering new avenues for tracking longitudinal disease dynamics in neuromuscular diagnostics. The developed automated feature extraction pipeline is integrated and available within the MUSCAT (MUSCle fAt Topology) library.
			\vspace{1.5em}
		\end{onecolabstract}
	}]
	
	\let\thefootnote\relax\footnotetext{\textit{This Work has not yet been peer-reviewed and is provided by the contributing Authors as a means to ensure timely dissemination of scholarly and technical Work on a noncommercial basis. Copyright and all rights therein are maintained by the Authors or by other copyright owners. It is understood that all persons copying this information will adhere to the terms and constraints invoked by each Author's copyright. This Work may not be reposted without explicit permission of the copyright owner.}}
	
	\section{Introduction}
	The clinical evaluation of muscular dystrophies relies heavily on Magnetic Resonance Imaging (MRI) to assess progressive muscle degeneration and weakness. Over time, functional muscle tissue (low intensity) is gradually replaced by pathological manifestations, which can be radiologically stratified into moderate fat replacement (intermediate intensity) and severe fat replacement (high intensity). Identifying the precise morphological patterns of this replacement is paramount, as many neuromuscular disorders (NMDs) present with overlapping symptoms but require fundamentally different therapeutic interventions.
	
	Historically, the evaluation of these scans has been guided by semi-quantitative visual grading systems, most notably the Mercuri scale. This foundational system successfully established a standard for radiological reporting and remains a cornerstone of clinical diagnostics. However, as the field moves towards early-stage interventions and gene therapies, there is a growing necessity to complement visual assessment with automated, continuous metrics. While traditional quantitative MRI (qMRI) techniques, such as Fat Fraction (FF) mapping—provide reproducible volumetric measurements, they compress the 3D complexity of the muscle architecture into a single scalar value. A muscle where 30\% of fat constitutes a solid mass and one where 30\% manifests as an aggressive, reticular "moth-eaten" network are clinically distinct, yet both generate identical FF values. Therefore, rather than seeking to replace FF, which remains a fundamental baseline metric, our Graph-Radiomics framework is designed to be highly complementary. It explicitly captures the 3D spatial heterogeneity that global FF mathematically dilutes.
	
	\section{State of the Art: Feature Extraction in Muscle MRI}
	
	\subsection{Visual Grading and Quantitative Volumetry}
	The Mercuri scale remains a milestone in standardizing radiological reporting. Nonetheless, it lacks the sensitivity required for sub-clinical disease monitoring due to its discrete, categorical nature \cite{carlier2016skeletal}. While Dixon-based Fat Fraction (FF) mapping overcomes inter-observer bias, it relies on global volumetry. As established in recent literature, relying exclusively on volumetric burden is a methodological limitation; FF is "blind" to the structural topology, failing to distinguish between benign focal fat accumulation and pathological, invasive networks. Furthermore, clinical practice often assumes symmetric progression. Mutations like DUX4 (FSHD) present with extreme, patchy asymmetry. Averaging standard volumetric scores from left and right limbs effectively masks these unique, genotypic signatures, highlighting the need for an advanced geometric profiling.
	
	\subsection{Radiomics and the "Bag-of-Pixels" Problem}
	Classical radiomics standardizes feature extraction using matrices like GLCM or GLSZM to identify textural changes. However, these methods operate on a "bag-of-pixels" paradigm. By aggregating pixels into global statistical matrices, they become spatially blind. They cannot distinguish between scattered "islands" of fat and a unified, invasive network. Consequently, due to the immense manual burden of 3D annotation, a vast majority of current clinical studies restrict their analysis to a single mid-thigh cross-section (e.g., at 33\% of the femur length) \cite{ogier2021overview, kemnitz2020clinical}. This "single-slice" paradigm intrinsically assumes longitudinal uniformity, severely hindering the evaluation of diseases that exhibit highly localized or progressing proximo-distal degeneration gradients. Comprehensive 3D spatial profiling is therefore mandatory to capture the true topological footprint of the pathology.
	
	\subsection{The Segmentation Bottleneck and Atlas Failure}
	The extraction of quantitative biomarkers intrinsically depends on accurate tissue segmentation. However, manual delineation is highly operator-dependent and impractically time-consuming for large 3D datasets. While automated atlas-based and active contour methods have shown efficacy in healthy cohorts, they often fail in advanced neuromuscular disorders \cite{ogier2021overview}. Extreme fatty infiltration and severe muscle atrophy obscure the fascia lata and erase natural inter-muscular boundaries, rendering predefined healthy atlases ineffective. This anatomical breakdown necessitates the use of adaptive, unsupervised segmentation frameworks that rely on probabilistic intensity distributions rather than rigid morphological priors, ensuring the extraction pipeline operates on accurately deconvoluted tissue masks.
	
	\subsection{Multi-Center Heterogeneity and Scale Information Leakage}
	The transition towards data-driven diagnostic models is further complicated by the heterogeneity of retrospective clinical annotations. As demonstrated by recent large-scale multi-center frameworks, aggregating visual scores from various institutions introduces differing quantization schemes (e.g., 4-point vs. 6-point discrete scales) \cite{verdu2025myoguide}. If specific rare diseases are predominantly annotated using a distinct scale within a historical cohort, machine learning models may inadvertently learn the grading scale artifact rather than the underlying pathophysiology, a phenomenon known as "scale information leakage." This highlights the critical need for automated pipelines that process continuous, image-driven spatial metrics directly, bypassing human quantization bias entirely.
	
	\subsection{The Deep Learning vs. Interpretability Gap}
	While Convolutional Neural Networks (CNNs) have achieved high classification accuracy, they face the "Data-Hunger" barrier in rare diseases, where expert-annotated 3D masks are practically unattainable \cite{verdu2025myoguide}. Furthermore, end-to-end CNNs often suffer from poor cross-scanner generalizability. Models trained on specific hardware configurations or field strengths (e.g., 1.5T vs. 3.0T) tend to overfit to hardware-specific signal intensities and contrast variations, leading to suboptimal performance on external datasets \cite{chen2025machine}. In contrast, explicitly extracted topological invariants, such as skeletal graphs, map the underlying physical geometry and connectivity of the tissue, offering a more robust and hardware-agnostic representation of the disease. In high-stakes diagnostics, clinicians require interpretable, physical metrics. Unlike "black-box" deep learning, our Graph-Radiomics framework is designed to yield biomarkers that are directly translatable to the biological reality of fascicular disruption.
	
	\subsection{The Tractography Limitation in Lipodegeneration}
	Historically, 3D structural analysis in muscle MRI has been strictly dominated by Diffusion Tensor Imaging (DTI) and tractography. DTI maps the orientation of healthy contractile muscle fibers by tracking the anisotropic diffusion of water molecules \cite{froeling2012diffusion}. While highly effective in healthy cohorts or sports medicine, DTI breaks down in advanced muscular dystrophies. As macroscopic fat replaces muscle tissue, the water signal diminishes, introducing massive partial volume effects and signal-to-noise (SNR) drops that render architectural tracking highly unreliable \cite{froeling2012diffusion}. Rather than struggling to map the surviving muscle fibers through heavy noise, our Graph-Radiomics framework reverses the diagnostic paradigm: it directly models and parameterizes the 3D architecture of the invasive fat network itself, remaining robust where DTI loses reliability.
	
	\subsection{Statistical vs. Morphological Graph-Radiomics}
	To overcome the limitations of conventional, spatially-blind radiomics, the broader medical imaging community is transitioning towards Graph-Radiomics. Recent breakthroughs in oncology have introduced graph-based descriptors (e.g., GrRAIL) to decode intra-tumoral heterogeneity \cite{battalapalli2025graph}. However, in oncological applications, graph nodes typically represent abstract mathematical clusters of statistical intensity, and edges represent generalized feature similarity. Their graphs form an abstract manifold of statistical heterogeneity. This paper introduces the translation of Graph-Radiomics to rare neuromuscular disorders via a uniquely \textit{morphological} approach. In our framework, the graph nodes and edges are not abstract statistical clusters, but direct representations of the physical skeletal infrastructure of fat infiltrates. This ensures that the extracted structural complexity directly mirrors the actual macroscopic anatomy, bridging the gap between abstract graph theory and clinically interpretable tissue morphology.
	
	\subsection{Statistical Rigor: Beyond P-Value}
	Finally, we address the common methodological pitfalls in radiomic discovery. Standard radiomic studies often extract hundreds of features and report only those with $p < 0.05$. However, according to modern statistical guidelines, $p$-values are highly sensitive to sample size and do not reflect the clinical magnitude of a finding. We employ the Epsilon-squared ($\epsilon^2$) effect size, as recommended in contemporary biostatistics \cite{tomczak2014need}, to ensure that our proposed features are not merely statistical noise, but reflect substantial pathological variations between genetic phenotypes.
	
	\section{Methods: Feature Engineering and Mathematical Basis}
	
	\subsection{Study Population and Tissue Segmentation}
	This study initially evaluated 1184 magnetic resonance imaging (MRI) scans acquired from patients enrolled in the multi-center CoMPaSS-NMD project. The analyzed cohort encompasses five distinct genetic neuromuscular phenotypes: DYSF (n = 458), CAPN3 (n = 343), GNE (n = 146), DMPK (n = 135), and DUX4 (n = 102). Prior to topological feature engineering, all MRI volumes underwent an automated intensity-based segmentation pipeline within the MUSCAT framework. 
	
	This unsupervised preprocessing step utilizes a Gaussian Mixture Model (GMM) to decompose the tissue intensity distributions. It robustly isolates the macroscopic functional muscle structure from pathological infiltration. The lipodegeneration is stratified into three distinct masks for comprehensive 3D profiling: Subtype 1 Fat (SF1), representing moderate fat replacement (intermediate intensity); Subtype 2 Fat (SF2), representing severe, macroscopic fat (high intensity); and General Fat (GF), which is the union of SF1 and SF2, representing the unified pathological burden. To capture the holistic disease footprint and ensure statistical robustness, features extracted from both the left and right legs were mathematically aggregated (averaged using the mean) and treated as a single unified profile per patient.
	
	The proposed feature extraction framework operates across five main architectural domains to holistically capture the 3D topology of the entire muscle volume. A rigorous correlation-based filter (Mutual Information) refined the initial feature set to 29 unique, orthogonal spatial biomarkers. The mathematical formulations below are uniformly applied to the GF, SF1, and SF2 masks independently, ensuring that the spatial representation is deterministic, reproducible, and hardware-agnostic. A comprehensive overview of the clinical and morphological interpretation of these five domains is provided in Table \ref{tab:clinical_interpretation}. Furthermore, the exact nomenclature and categorization of all extracted variables across the GF, SF1, and SF2 masks are explicitly detailed in the feature dictionary (Table \ref{tab:feature_dictionary}).
	
	\begin{table*}[t]
		\centering
		\caption{Clinical Translation of Engineered 3D Graph-Radiomic Biomarkers}
		\label{tab:clinical_interpretation}
		\renewcommand{\arraystretch}{1.5}
		\normalsize 
		\begin{tabular}{p{0.25\linewidth} p{0.35\linewidth} p{0.35\linewidth}}
			\hline
			\textbf{Topological Characteristic} & \textbf{Morphological / Radiological Pattern} & \textbf{Clinical Interpretation in NMDs} \\ \hline
			\textbf{Topological Complexity} \cite{blum1967transformation, battalapalli2025graph, verdu2025myoguide} \newline \textit{(Skel\_Nodes \& Skel\_Cycles)} & Measures the density of branching intersections and closed loops within the fat skeleton. & High values indicate an aggressive, reticular ("web-like") inter-fascicular permeation, heavily disrupting contractile geometry. Explicitly identifies the highly branched networks often seen in severe DUX4 phenotypes. \\ 
			\textbf{Fragmentation} \cite{cai2019, mercuri2005, verdu2025myoguide} \newline \textit{(Blob\_Count \& Size)} & Quantifies the absolute number and median size of disconnected, discrete spatial fat islands (Connected Components). & A high count combined with small fragment size reflects a scattered, multi-focal "moth-eaten" pattern. This translates subjective visual grading into a continuous metric for early, patchy lipodegeneration. \\
			\textbf{Spatial Vectoring} \cite{hu1962visual, verdu2025myoguide} \newline \textit{(Center of Mass (CoM))} & Maps the exact 3D center of gravity of the pathology within the anatomical compartment. & Provides objective coordinates to distinguish specific asymmetrical involvement (e.g., anterior vs. posterior dominance), tracking spatial disease progression vectors explicitly. \\
			\textbf{Interface Dynamics} \cite{carlier2016skeletal, verdu2025myoguide} \newline \textit{(SF2\_To\_SF1\_Dist\_Min)} & Calculates the shortest Euclidean distance between severe fat replacement (SF2) and moderate fat replacement (SF1) masks. & Evaluates the synergistic growth at the pathological tissue interface. Extremely low distances indicate densely intertwined degenerative processes, separating LGMD variants (CAPN3 vs. DYSF). \\ 
			\textbf{Geometric Orientation} \cite{cai2019, verdu2025myoguide} \newline \textit{(MainAxis\_Z \& Flatness)} & Uses PCA eigenvectors to determine the primary axis of spatial variance and shape stretch. & Determines if fat infiltration propagates as a continuous longitudinal cylinder or spreads horizontally across inter-muscular planes. This geometric profiling highlights the necessity of full 3D mapping. \\ \hline
		\end{tabular}
	\end{table*}
	
	\begin{table*}[t]
		\centering
		\caption{Feature Dictionary: Explicit categorization of the engineered 3D spatial biomarkers by architectural domain. With the exception of Interface Dynamics and overall muscle percentage, features are systematically extracted for the General Fat (GF), Subtype 1 (SF1), and Subtype 2 (SF2) masks.}
		\label{tab:feature_dictionary}
		\renewcommand{\arraystretch}{1.6}
		\normalsize 
		\begin{tabular}{>{\raggedright\arraybackslash}p{0.25\linewidth} >{\raggedright\arraybackslash}p{0.70\linewidth}}
			\hline
			\textbf{Architectural Domain} & \textbf{Engineered Biomarkers (Exact Nomenclature)} \\ \hline
			\textbf{Topological Complexity} & \texttt{\_Skel\_Nodes}, \texttt{\_Skel\_Cycles}, \texttt{\_Skel\_Components}, \texttt{\_Skel\_Len\_Total} \newline \footnotesize{(Each calculated independently for GF, SF1, and SF2)} \\ 
			\textbf{Fragmentation \& Volumetry} & \texttt{\_Blob\_Count}, \texttt{\_Global\_Pct}, \texttt{\_BlobSize\_Mean}, \texttt{\_BlobSize\_Median}, \texttt{\_BlobSize\_Std}, \texttt{\_BlobSize\_Max} \newline \footnotesize{(Each calculated independently for GF, SF1, and SF2)}, \texttt{Muscle\_Global\_Pct} \\
			\textbf{Spatial Vectoring \& Distribution} & \texttt{\_CoM\_X}, \texttt{\_CoM\_Y}, \texttt{\_CoM\_Z}, \texttt{\_Depth\_Mean}, \texttt{\_Depth\_Median}, \texttt{\_Depth\_Std}, \texttt{\_Depth\_Max} \newline \footnotesize{(Each calculated independently for GF, SF1, and SF2)} \\
			\textbf{Geometric Orientation} & \texttt{\_Elongation}, \texttt{\_Flatness}, \texttt{\_MainAxis\_Z} \newline \footnotesize{(Each calculated independently for GF, SF1, and SF2)} \\
			\textbf{Interface Dynamics} & \texttt{SF2\_To\_SF1\_Dist\_Min}, \texttt{SF2\_To\_SF1\_Dist\_Mean}, \texttt{SF2\_To\_SF1\_Contact\_Ratio} \newline \footnotesize{(Calculated exclusively between the SF2 and SF1 tissue boundaries)} \\ \hline
		\end{tabular}
	\end{table*}
	
	\subsection{Volumetric Macro-structure and Morphological Fragmentation}
	Basic volumetrics are initially derived from voxel counting multiplied by 3D spatial spacing to obtain the absolute volumetric extent ($V_{total}$) in mm$^3$:
	$$V_{total} = N \times (s_x \cdot s_y \cdot s_z)$$
	where $s_x, s_y, s_z$ represent the physical voxel spacing of the MRI scan (in millimeters). Crucially, to eliminate anthropometric bias (i.e., natural variations in patient height and limb size), this absolute volume is strictly not used as a standalone biomarker. Instead, it is normalized against the total anatomical compartment volume, yielding the relative disease burden (\textit{\_Global\_Pct}). 
	
	Fragmentation is calculated using 3D Connected Component Analysis based on 26-connectivity to identify discrete, non-overlapping spatial structures (islands) $C = \{C_1, C_2, \dots, C_k\}$, implemented computationally via the \textit{scikit-image} library \cite{van2014scikit}. The 26-neighborhood criterion was deliberately chosen over 6- or 18-connectivity to prevent artificial computational fragmentation of thin, diagonally oriented fat infiltrates that spread across inter-muscular fasciae.
	
	\subsection{Spatial Distribution and Vectoring}
	These features are based on exact Euclidean distance transforms and the classical theory of 3D image moments \cite{hu1962visual} for calculating geometric centroids. The Normalized Center of Mass ($CoM$) provides spatial coordinates tracking the absolute center of gravity of the pathology:
	$$CoM_x = \frac{1}{N} \sum_{i \in \Omega} i_x \cdot V(i)$$
	where $\Omega$ represents the segmented tissue region, $N$ is the total number of voxels within the mask, $i_x$ is the specific coordinate, and $V(i)$ is the binary mask value (1 or 0) at that coordinate. Similar equations apply to the Y and Z axes.
	
	\subsection{Geometric Shape and Orientation}
	Evaluated using Principal Component Analysis (PCA) on the voxel coordinates to extract the spatial covariance matrix describing spatial variance. By calculating the orthogonal eigenvalues ($\lambda_1 \ge \lambda_2 \ge \lambda_3$), we quantify the macroscopic geometric shape. \textit{Elongation} is defined as the square root of the ratio of the two largest eigenvalues:
	$$Elongation = \sqrt{\frac{\lambda_2}{\lambda_1}}$$
	Similarly, \textit{Flatness} is defined as $\sqrt{\lambda_3 / \lambda_2}$. 
	
	\subsection{Interface Dynamics Between Fat Replacement Stages}
	Calculated using 3D spatial dilation and distance mapping to evaluate the physical intersection and proximity between the severe fat replacement mask ($F_2$) and the moderate fat replacement mask ($F_1$). The minimum interface distance is calculated as:
	$$D_{min} = \min_{p \in F_2, q \in F_1} \|p - q\|_2$$
	where $p$ and $q$ represent the spatial coordinates belonging to the respective tissue classes, and $\|p - q\|_2$ denotes the Euclidean distance between their boundaries. 
	
	\subsection{3D Graph-based Topology (Network Architecture)}
	Derived from the Medial Axis Transform (MAT) theory initially proposed by Blum \cite{blum1967transformation}, the 3D volumetric fat mask undergoes morphological skeletonization. This complex morphological reduction is computationally executed producing a 1-voxel-thick medial line while strictly preserving the underlying topology.
	
	Once skeletonized, the fat infrastructure is formally mapped into a mathematical undirected graph $G=(V,E)$, where $V$ is the set of topological vertices (branching nodes and terminal endpoints) and $E$ represents the set of connecting skeletal edges. The vertices $V$ are defined by utilizing a continuous Coordinate k-D Tree to query voxel neighborhoods within a spatial radius of $r \le 1.8$ voxels. 
	
	Crucially, the Total Network Length (\textit{Skel\_Len\_Total}) is computed by multiplying the cumulative length of all edges by the mean spatial spacing of the MRI scan. Furthermore, the Number of Closed Loops (\textit{Skel\_Cycles}) is mathematically quantified using the cyclomatic complexity formula:
	$$C = |E| - |V| + N_{comp}$$
	where $|E|$ is the total number of connected skeletal edges, $|V|$ is the number of branch/terminal vertices, and $N_{comp}$ signifies the number of fully disconnected spatial sub-graphs. 
	
	\begin{figure*}[t]
		\centering
		\begin{subfigure}[b]{0.32\textwidth}
			\centering
			\includegraphics[width=\textwidth]{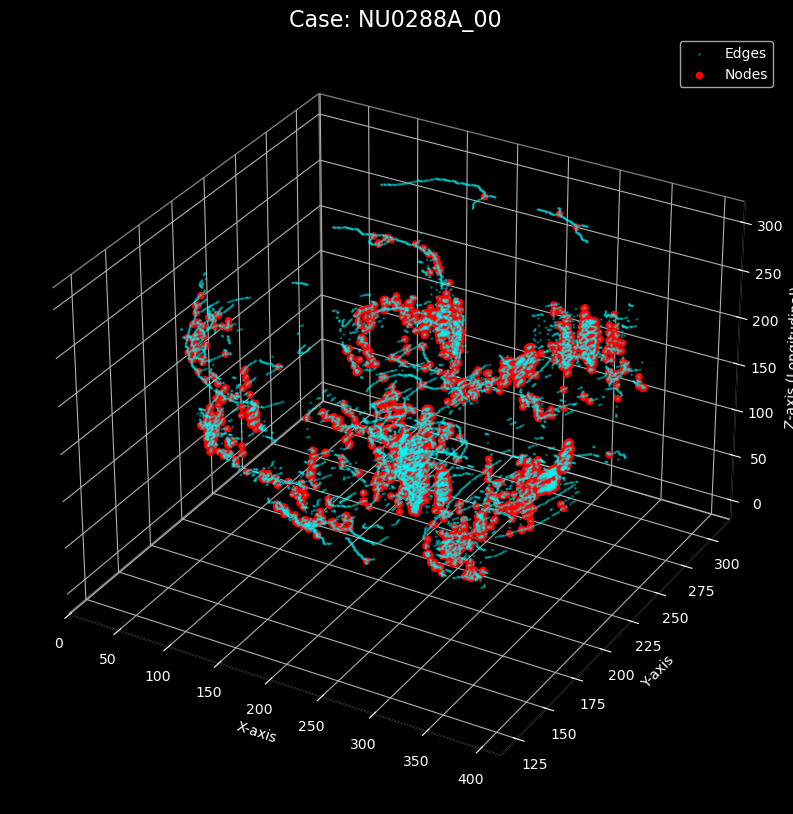}
			\caption{Sparse/Fragmented Pattern}
			\label{fig:graph1}
		\end{subfigure}
		\hfill
		\begin{subfigure}[b]{0.32\textwidth}
			\centering
			\includegraphics[width=\textwidth]{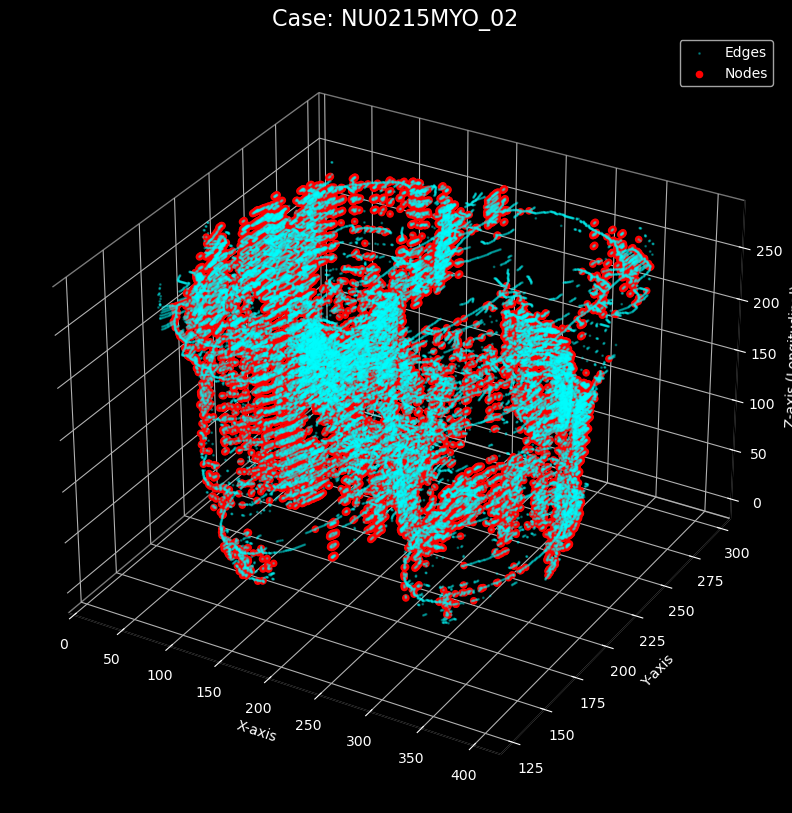}
			\caption{Structured Linear Streaks}
			\label{fig:graph2}
		\end{subfigure}
		\hfill
		\begin{subfigure}[b]{0.32\textwidth}
			\centering
			\includegraphics[width=\textwidth]{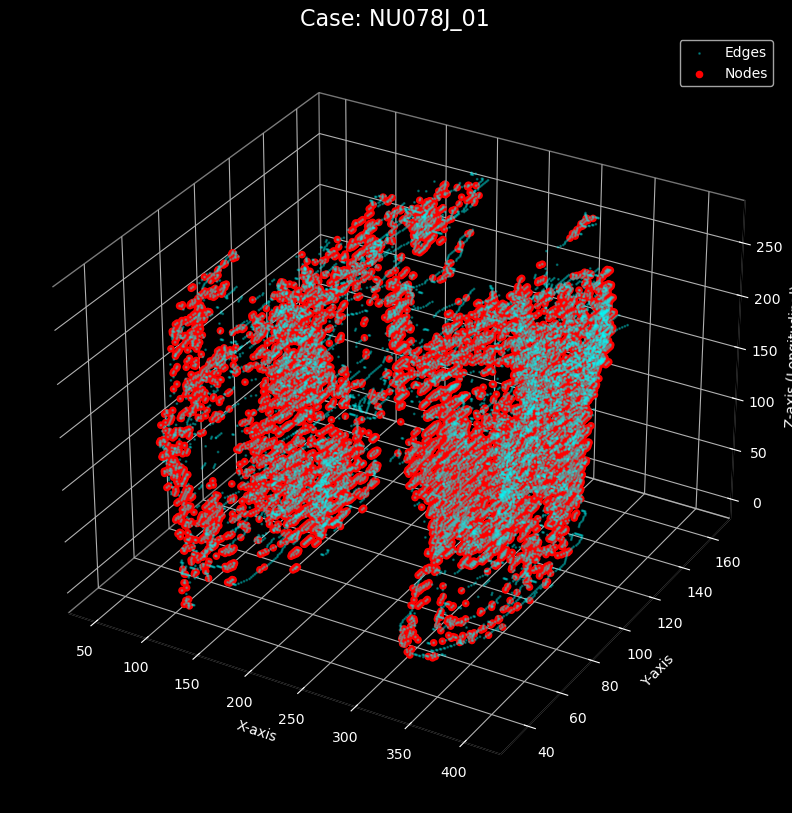}
			\caption{Complex Reticular Web}
			\label{fig:graph3}
		\end{subfigure}
		\caption{Visual representation of the extracted 3D Graph-Radiomics topology from macroscopic lipodegeneration masks. The original volumetric fat segments are reduced to 1-voxel-thick medial axis skeletons, represented by connecting paths (cyan edges), with computationally extracted branching intersections and terminal points (red nodes). These 3D spatial networks allow for the exact mathematical quantification of structural complexity, such as the density of inter-fascicular permeation, which standard volumetric methods fail to capture.}
		\label{fig:graphs}
	\end{figure*}

	\section{Results}
	
	\subsection{Global Variance and Effect Size}
	The statistical screening of the engineered biomarkers was conducted using the non-parametric Kruskal-Wallis H-test \cite{kruskal1952use}. This method securely evaluated the global variance, testing the alternative hypothesis that at least one disease group differs significantly from the rest. 
	
	To explicitly quantify the magnitude of differentiation between the genetic phenotypes, independent of sample size bias, the Epsilon-squared ($\epsilon^2$) effect size was calculated \cite{tomczak2014need}. The top 10 discriminative global features demonstrating the highest statistical power are reported in Table \ref{tab:kw_results}. Notably, topological network features such as \textit{SF1\_Skel\_Nodes} achieved a large effect size of $\epsilon^2 = 0.2656$, and \textit{GF\_Skel\_Nodes} achieved $\epsilon^2 = 0.2542$, indicating profound differentiation based on the 3D graph complexity of the pathology rather than simple volumetric measures.
	
	\begin{table}[h]
		\centering
		\caption{Top 10 discriminative engineered biomarkers identified by the Kruskal-Wallis H-test, sorted by $\epsilon^2$ effect size. A comprehensive evaluation is provided in Supplementary Table S1.}
		\label{tab:kw_results}
		\setlength{\tabcolsep}{3pt}
		\begin{tabular}{@{}p{2.1cm} >{\raggedright\arraybackslash}p{3.2cm} c c@{}}
			\hline
			\textbf{Category} & \textbf{Topological Characteristic} & \textbf{H-Stat} & \textbf{$\epsilon^2$} \\ \hline
			Topology & SF1\_Skel\_Nodes & 314.17 & 0.2656 \\
			Topology & GF\_Skel\_Nodes & 300.68 & 0.2542 \\
			Fragmentation & SF2\_BlobSize\_ \newline Median & 299.81 & 0.2534 \\
			Fragmentation & GF\_BlobSize\_ \newline Median & 277.95 & 0.2349 \\
			Interface & SF2\_To\_SF1\_ \newline Dist\_Min & 247.44 & 0.2092 \\
			Topology & GF\_Skel\_ \newline Components & 222.73 & 0.1883 \\
			Fragmentation & SF1\_BlobSize\_ \newline Median & 219.20 & 0.1853 \\
			Topology & SF2\_Skel\_ \newline Cycles & 184.82 & 0.1562 \\
			Spatial Vect. & SF1\_CoM\_Z & 159.87 & 0.1351 \\
			Fragmentation & SF1\_BlobSize\_ \newline Max & 155.51 & 0.1315 \\ \hline
		\end{tabular}
	\end{table}
	
	\subsection{Pairwise Post-Hoc Analysis}
\begin{figure*}[t]
		\centering
		\begin{minipage}[c]{0.88\textwidth}
			\begin{subfigure}[b]{0.32\linewidth}
				\centering
				\includegraphics[width=\textwidth]{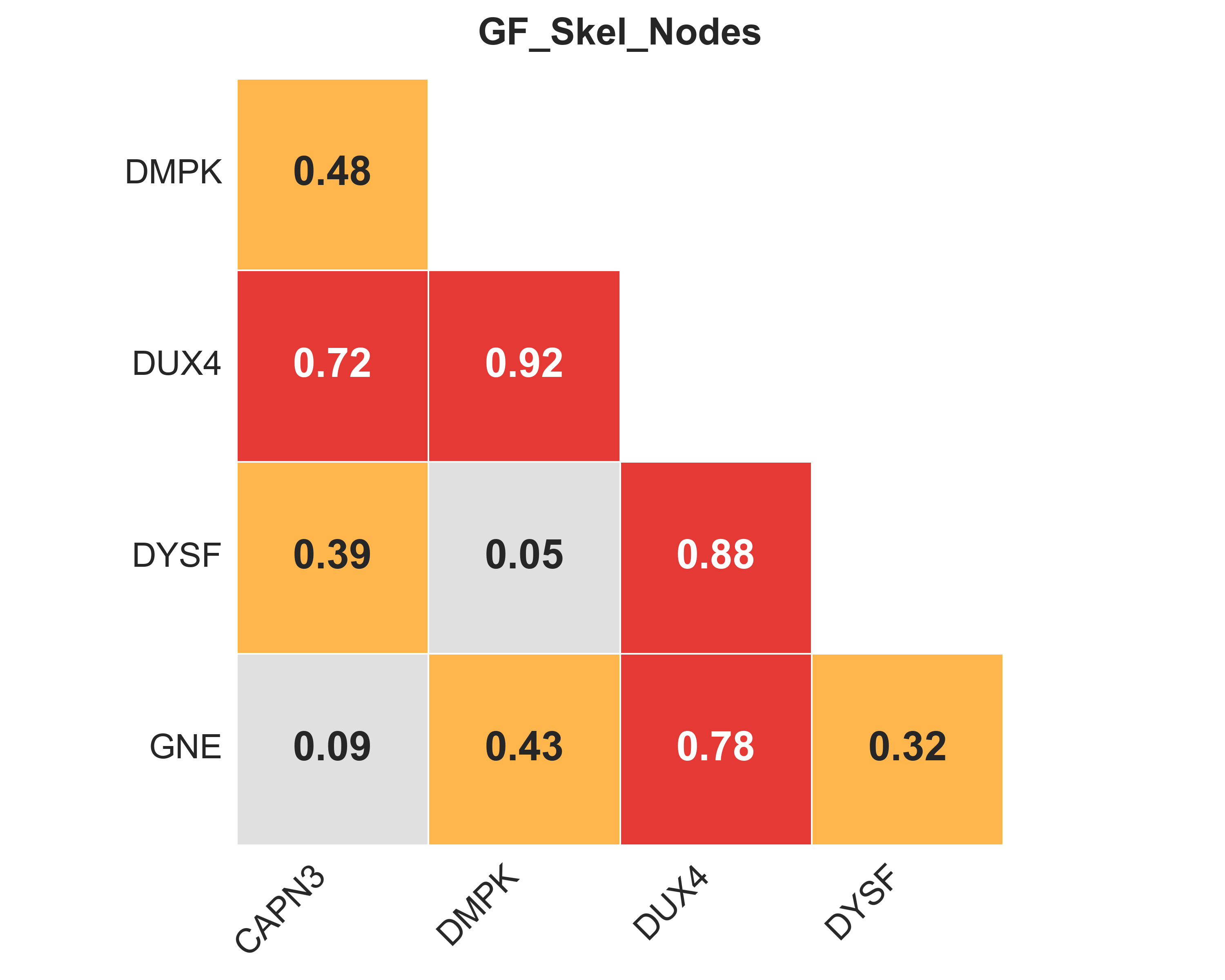}
				\caption{General Fat Topological Nodes}
			\end{subfigure}
			\hfill
			\begin{subfigure}[b]{0.32\linewidth}
				\centering
				\includegraphics[width=\textwidth]{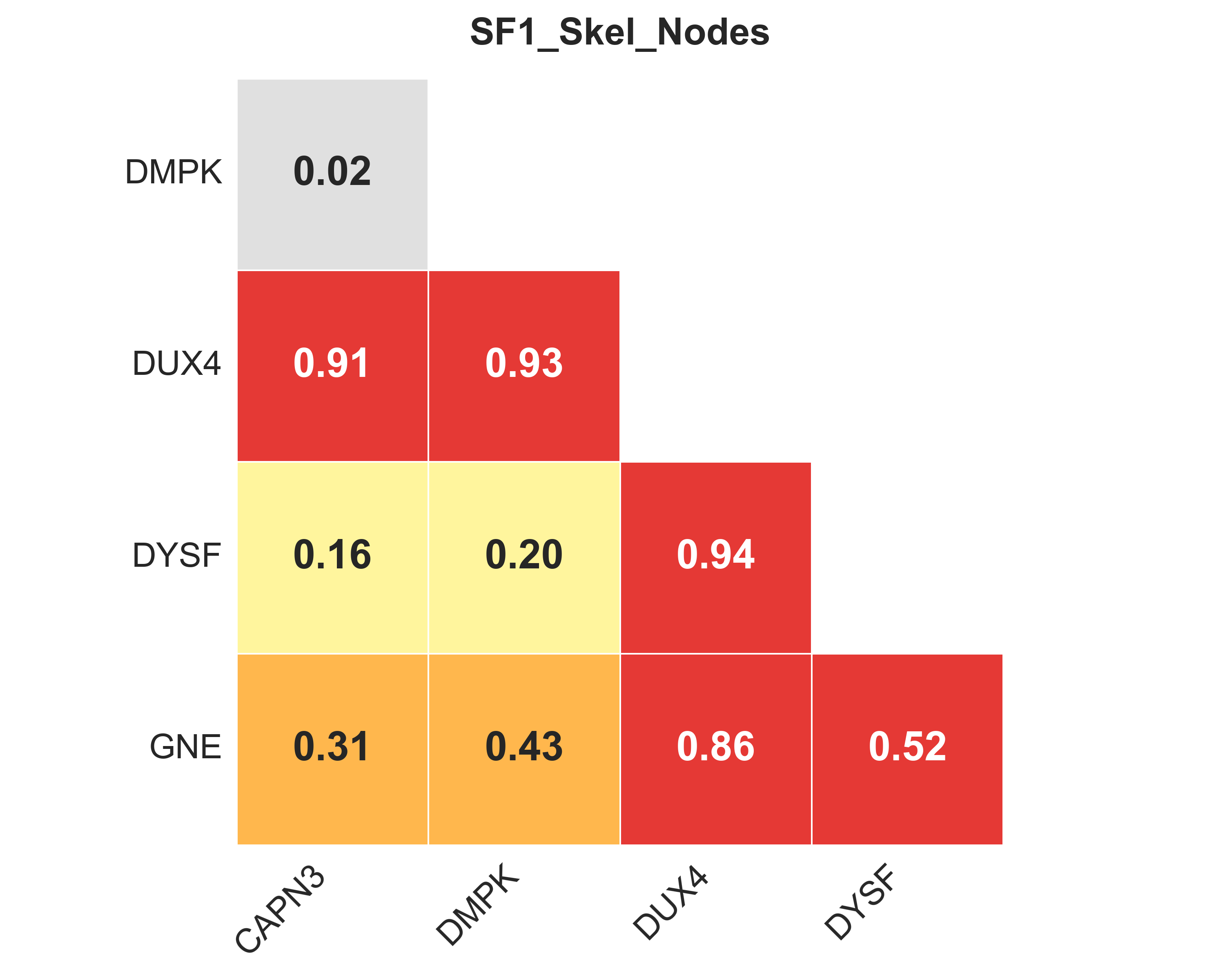}
				\caption{Moderate Fat Topological Nodes}
			\end{subfigure}
			\hfill
			\begin{subfigure}[b]{0.32\linewidth}
				\centering
				\includegraphics[width=\textwidth]{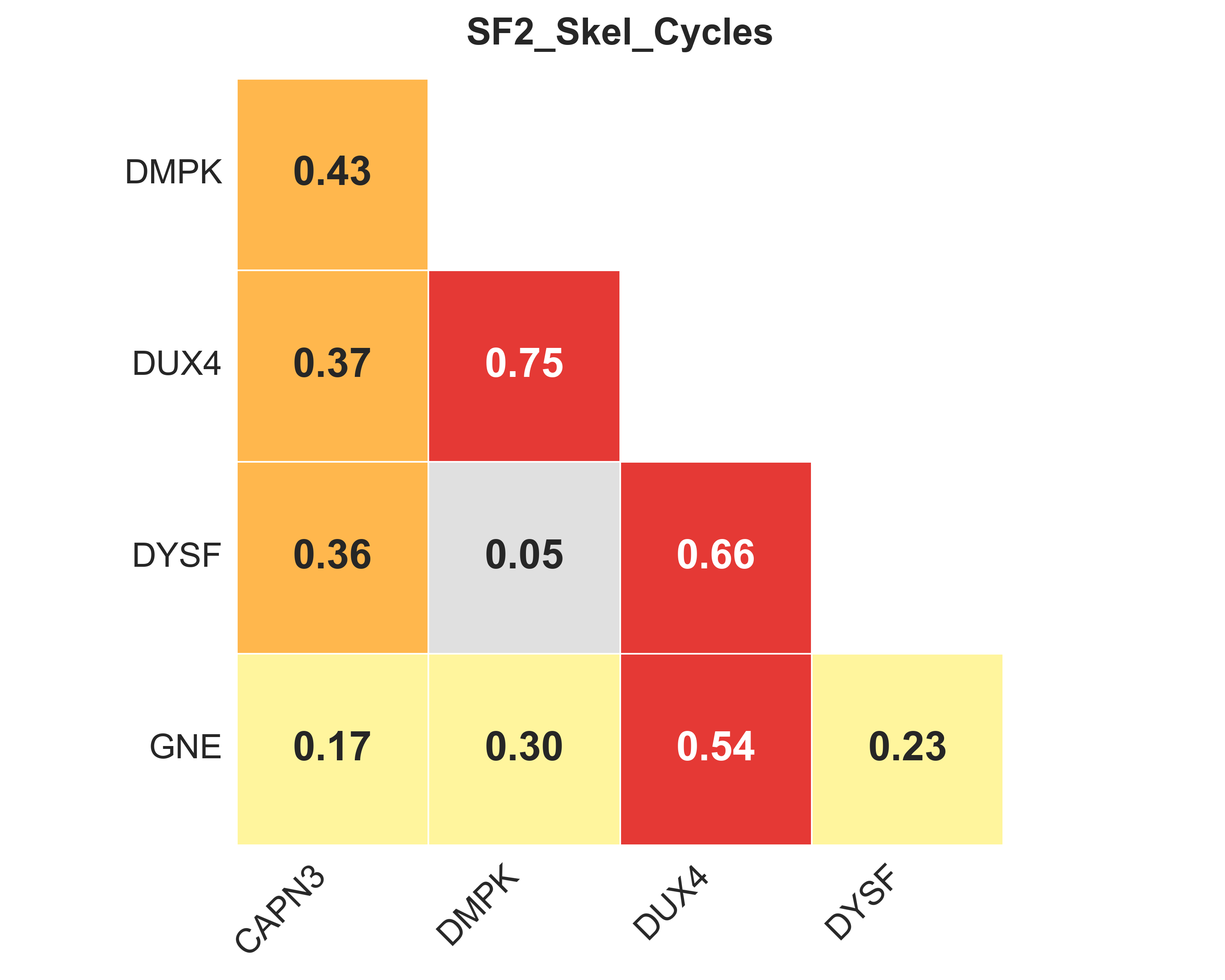}
				\caption{Severe Fat Skeletal Cycles}
			\end{subfigure}
			
			\vspace{1.5em}
			
			\begin{subfigure}[b]{0.32\linewidth}
				\centering
				\includegraphics[width=\textwidth]{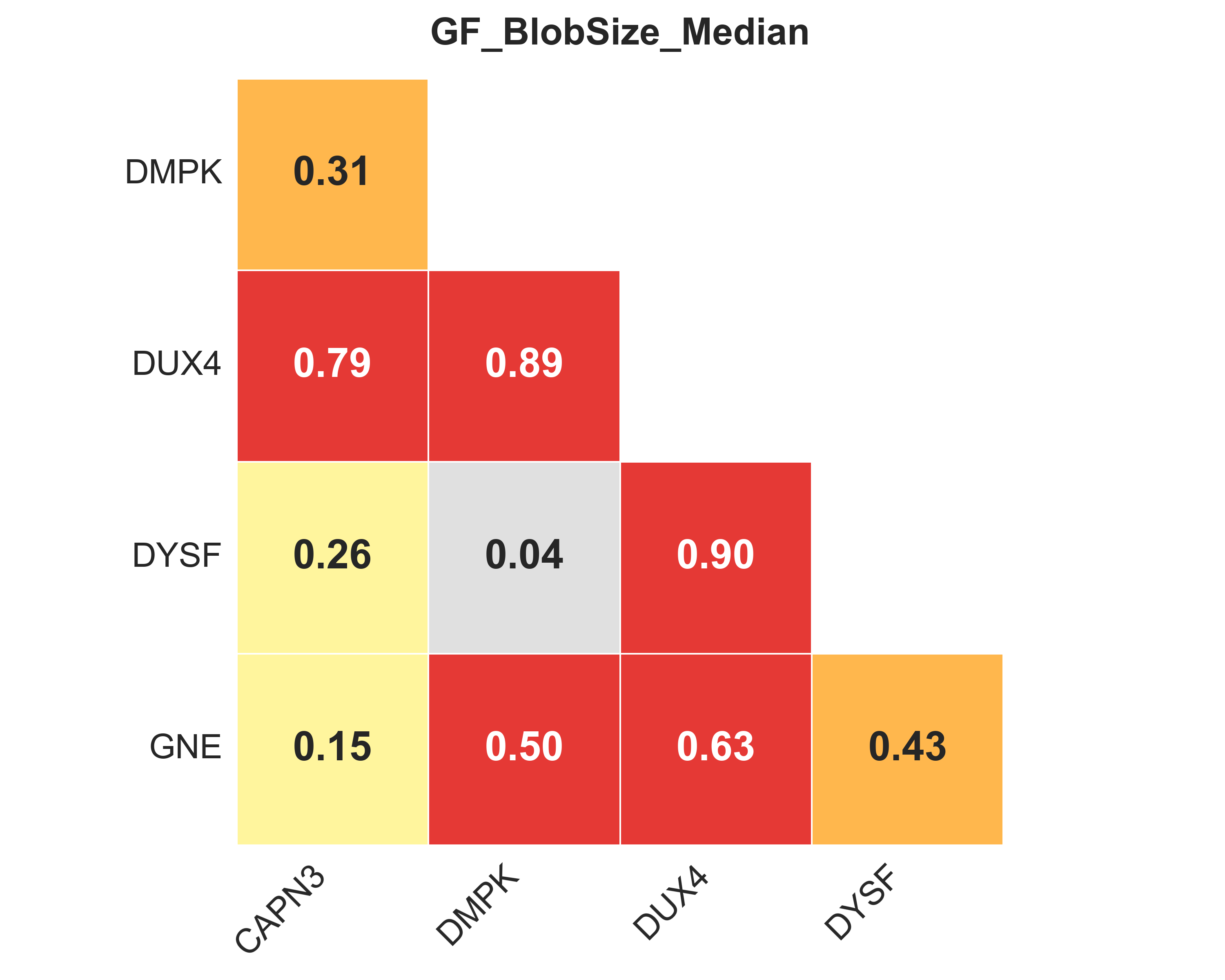}
				\caption{General Fat Median Fragment Size}
			\end{subfigure}
			\hfill
			\begin{subfigure}[b]{0.32\linewidth}
				\centering
				\includegraphics[width=\textwidth]{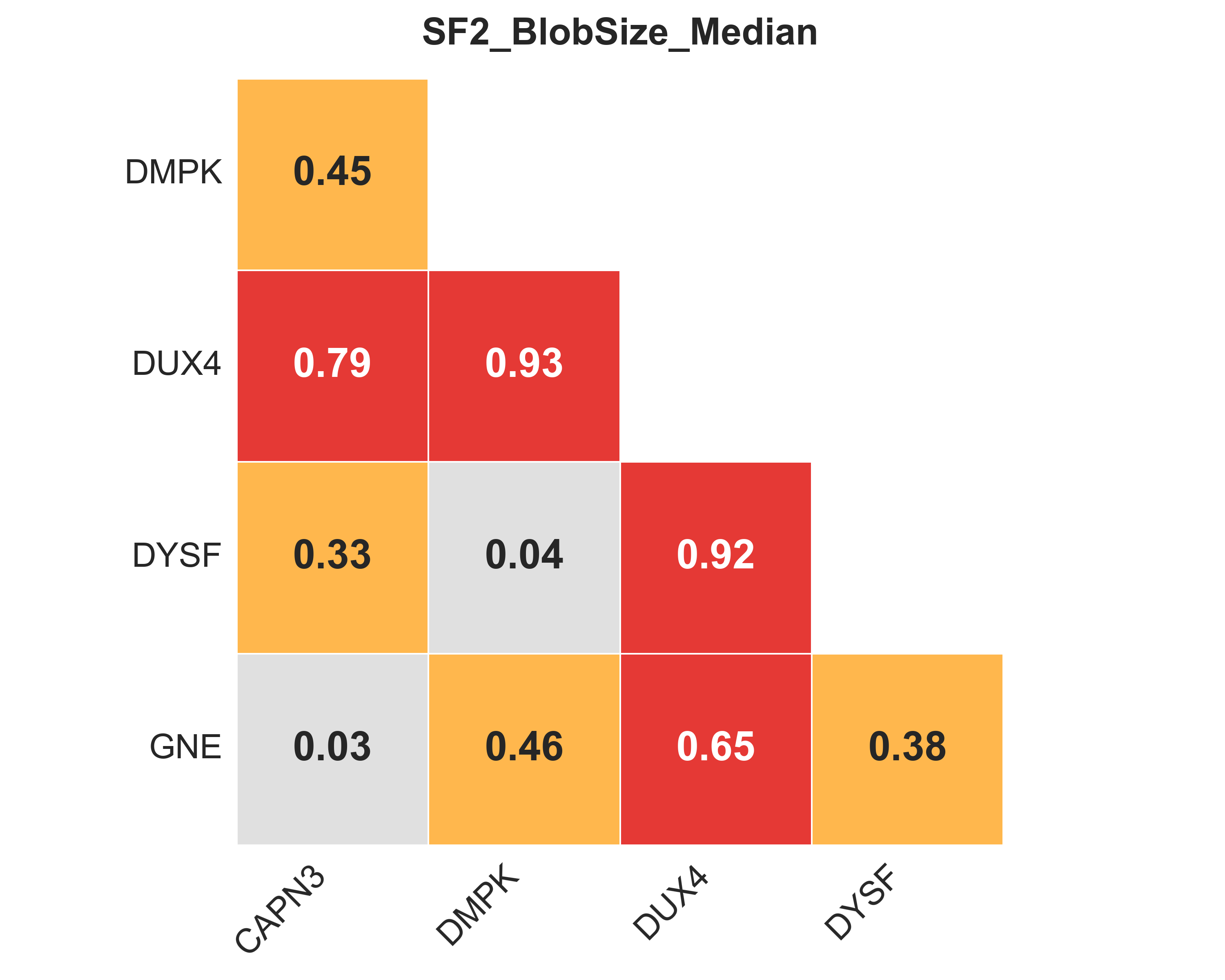}
				\caption{Severe Fat Median Fragment Size}
			\end{subfigure}
			\hfill
			\begin{subfigure}[b]{0.32\linewidth}
				\centering
				\includegraphics[width=\textwidth]{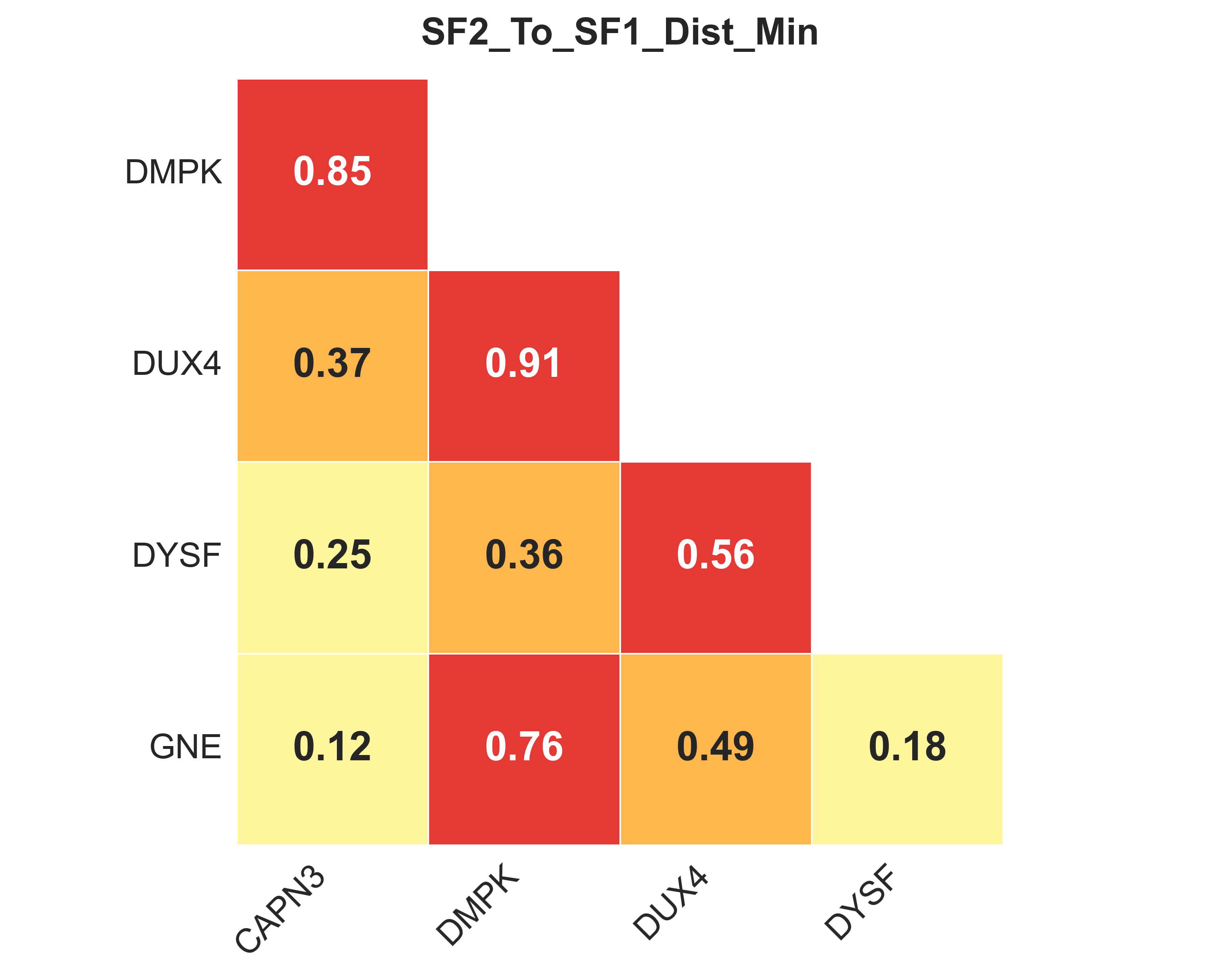}
				\caption{Severe-to-Moderate Fat Interface}
			\end{subfigure}
		\end{minipage}%
		\hfill
		\begin{minipage}[c]{0.1\textwidth}
			\centering
			\includegraphics[height=5.5cm, keepaspectratio]{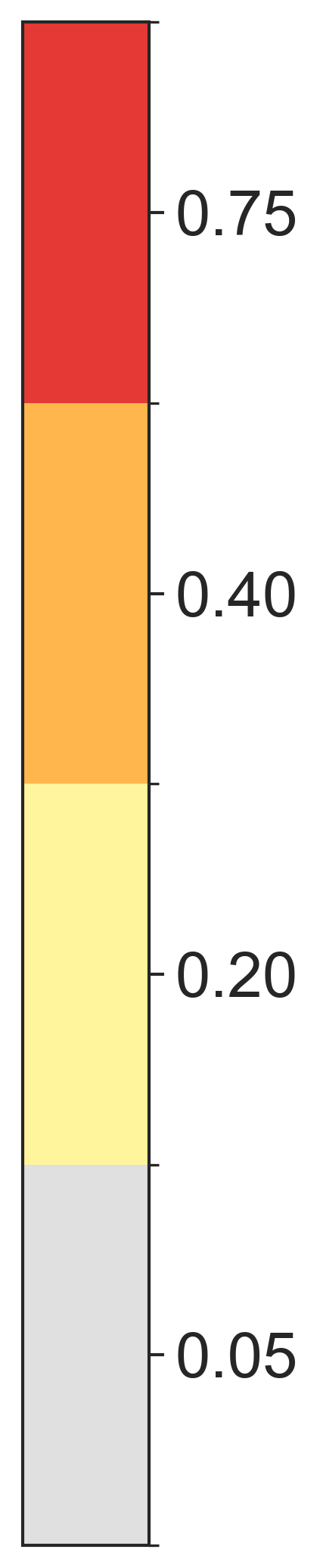} 
		\end{minipage}
		
		\caption{Pairwise effect size heatmaps derived from Dunn's post-hoc analysis across the top-performing global descriptors. The matrices visualize the magnitude of differentiation between the neuromuscular phenotypes. The accompanying color bar on the right defines the standardized effect size scale (r).}
		\label{fig:heatmaps}
	\end{figure*}

	To identify local differences in rank sums between specific disease subgroups, Dunn’s Post-Hoc tests \cite{dunn1964multiple} were executed. The pairwise evaluations demonstrated the capability of the topological and geometric features to isolate specific phenotypes based on their entire 3D volume. For instance, the global topological interface distance (\textit{SF2\_To\_SF1\_Dist\_Min}) successfully isolated specific phenotypes, indicating varying degrees of intertwining between moderate and severe fat. Furthermore, topological descriptors such as \textit{GF\_Skel\_Nodes} and \textit{SF1\_Skel\_Nodes} yielded large pairwise effect sizes (e.g., isolating DUX4 from DMPK and DYSF with effect sizes $r > 0.88$). A comprehensive visual representation of these pairwise magnitudes is presented in Figure \ref{fig:heatmaps}. 
	
	Detailed descriptive statistics (including mean, median, standard deviation, and quantiles) along with comprehensive Kruskal-Wallis findings for all engineered biomarkers are provided in Supplementary Table S1. Furthermore, the complete pairwise effect size matrices derived from Dunn's post-hoc analysis for all genetic phenotypes are accessible in Supplementary Table S2.
	
	\subsection{UMAP Dimensionality Reduction}
\begin{figure}[h]
		\centering
		\includegraphics[width=\linewidth]{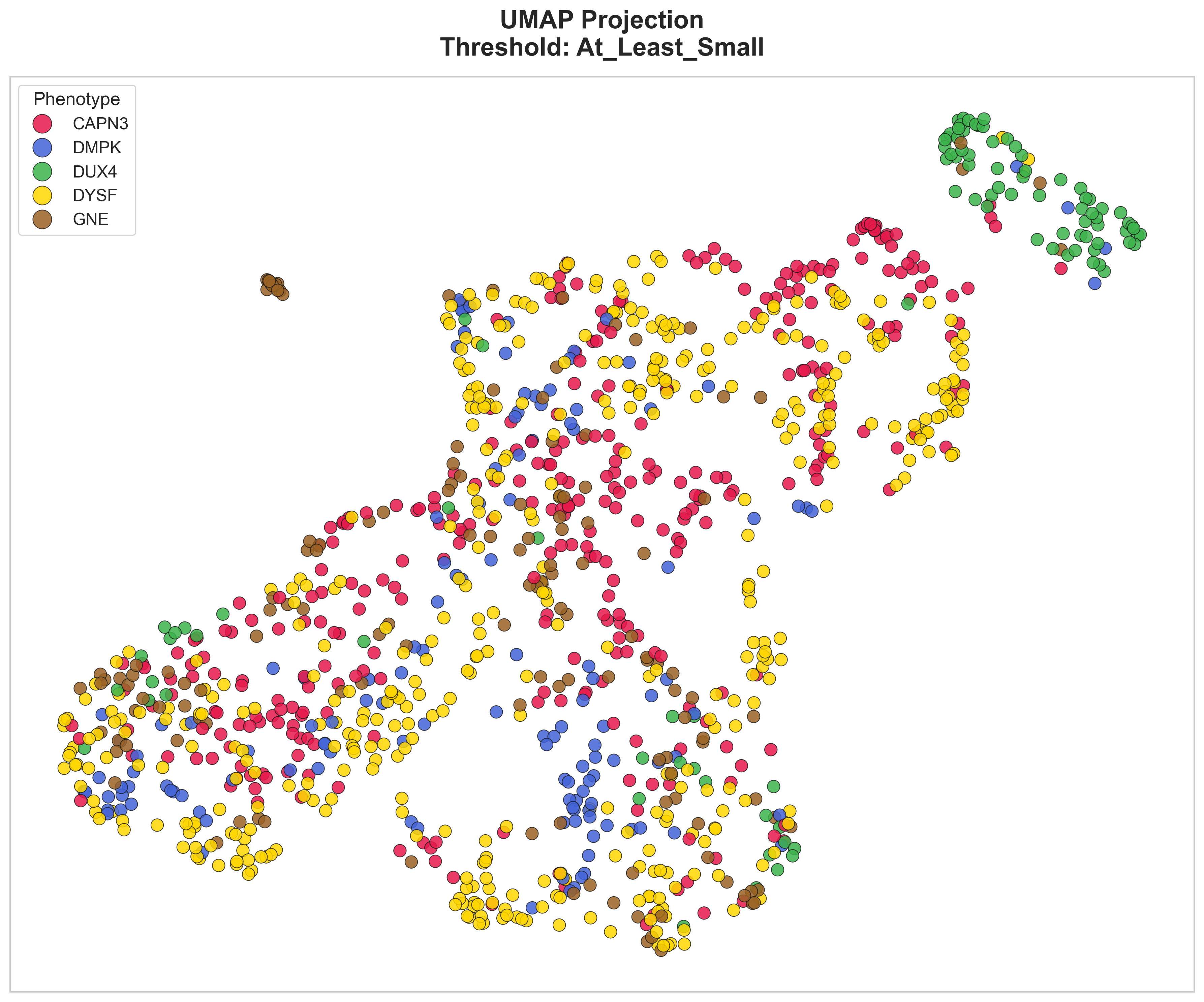} 
		\caption{UMAP Projection of the patient cohort based on the global biomarkers exhibiting at least a small effect size ($\epsilon^2 \ge 0.01$), pre-processed with PCA to retain 95\% of spatial variance. The visualization confirms the capability of this rigorously selected 3D feature subset to differentiate genetic phenotypes based on their entire volume topology, notably isolating DUX4 and GNE variants into distinct spatial clusters.}
		\label{fig:umap}
	\end{figure}

	To evaluate the holistic, multi-dimensional discriminative potential of the engineered global features, an unsupervised Uniform Manifold Approximation and Projection (UMAP) \cite{mcinnes2018umap} was applied to project the high-dimensional feature space into a 2D manifold while preserving local and global data structures. All spatial biomarkers demonstrating at least a small global effect size ($\epsilon^2 \ge 0.01$, yielding 28 features) were utilized. To prevent the curse of dimensionality and filter out statistical noise, this high-dimensional profile was first reduced via Principal Component Analysis (PCA, retaining 95\% of variance) before being projected into a 2D manifold without any exposure to the genetic labels (Figure \ref{fig:umap}).
	
\begin{figure}[h]
		\centering
		\includegraphics[width=\linewidth]{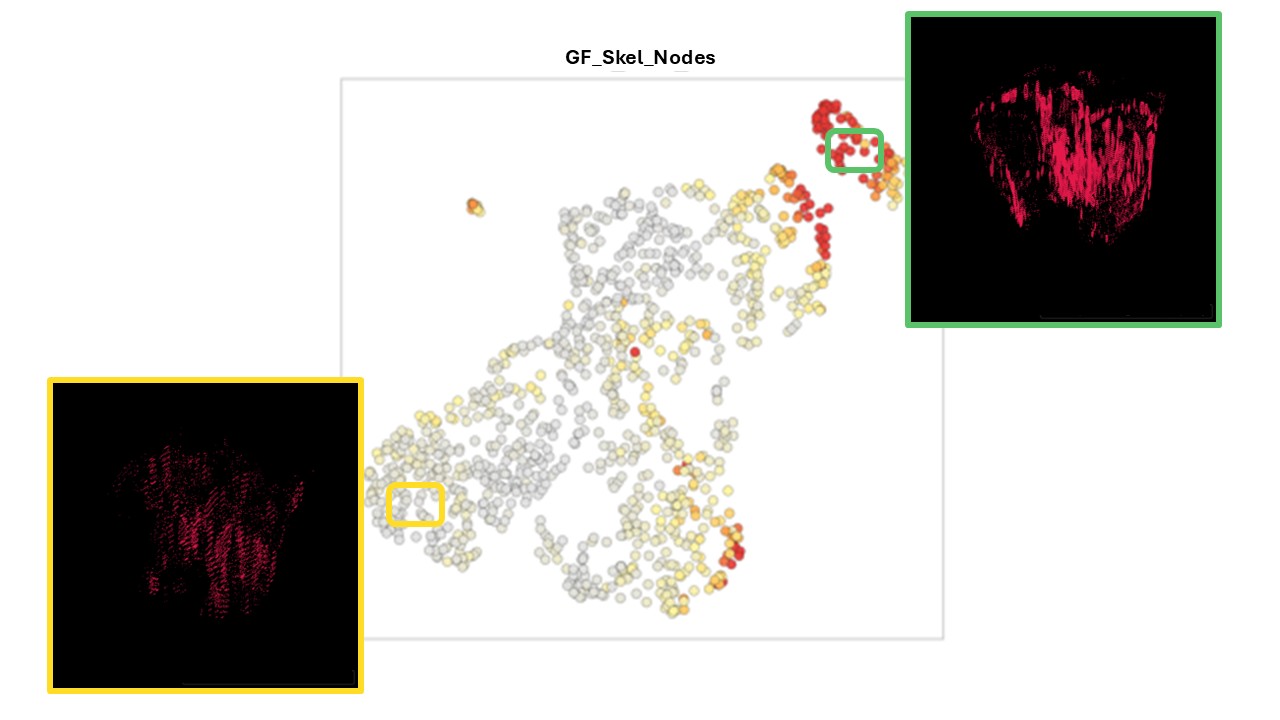}
		\caption{Visual validation of the 3D topological feature space. The central UMAP scatter plot displays the gradient for skeletal complexity (\textit{GF\_Skel\_Nodes}). The green frame illustrates the physical 3D topological skeleton of a DUX4 patient from the isolated high-complexity cluster, demonstrating a dense, highly branched network. The yellow frame shows a contrasting DYSF patient from the main continuous cluster, exhibiting lower topological complexity with linear fat propagation.}
		\label{fig:dux4_diff}
	\end{figure}

	To visually validate the spatial separation observed on the UMAP manifold, specifically the isolation of the DUX4 cohort, a direct structural comparison was performed using the \textit{GF\_Skel\_Nodes} gradient (Figure \ref{fig:dux4_diff}). The projection maps the 2D embedding directly to the physical 3D architecture of the extracted tissue. Patients in the isolated DUX4 cluster (green frame) exhibit an extreme density of topological branching nodes, forming a complex, multi-directional network. In contrast, patients located in the main continuous LGMD cluster, such as DYSF variants (yellow frame), present with significantly fewer branching nodes, characterized by parallel, longitudinal fat propagation.
	
\begin{figure*}[h]
		\centering
		\includegraphics[width=\textwidth]{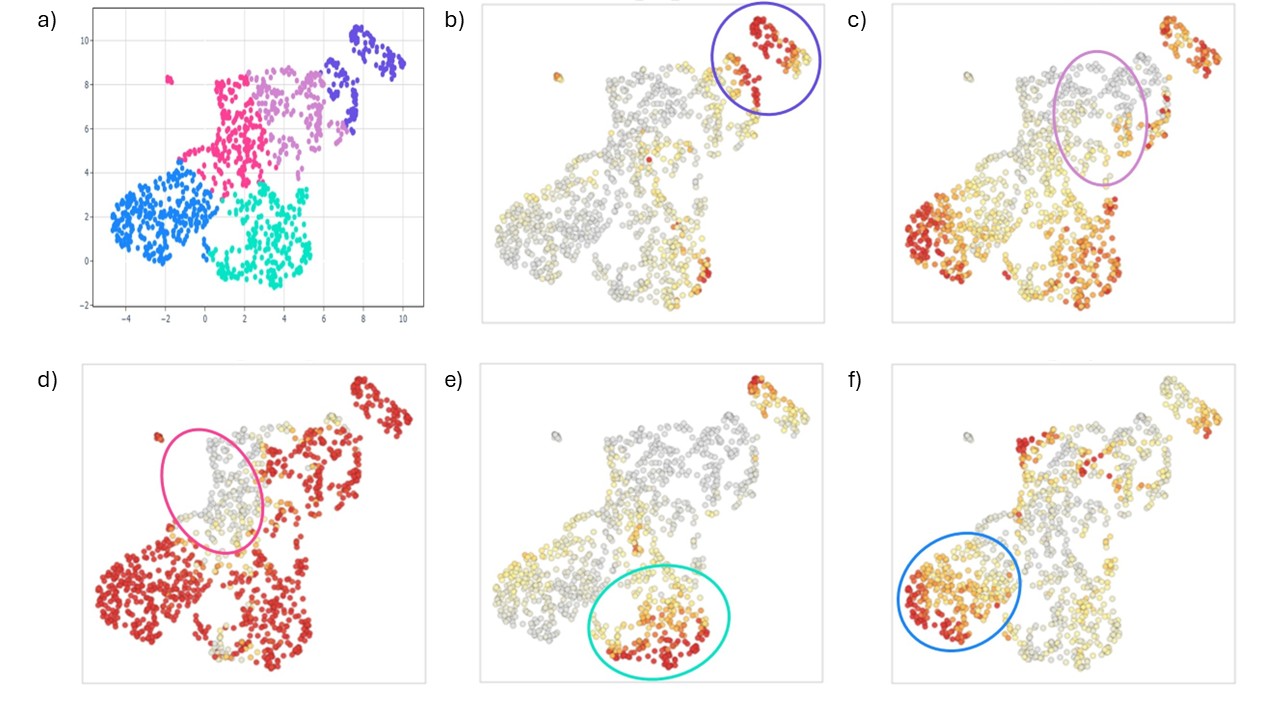} 
		\caption{Multi-dimensional UMAP projection and feature gradient mapping of the patient cohort. (a) Unsupervised PARC clustering identifying five distinct topological groups based on whole-muscle 3D spatial features. (b-f) Gradient maps of individual spatial biomarkers highlighting the structural signatures driving the manifold topology: (b) \textit{GF\_Skel\_Nodes} isolating the highly branched dark blue cluster; (c) \textit{SF1\_CoM\_Z} distinguishing the proximodistal gradient in the violet cluster; (d) \textit{GF\_MainAxis\_Z} identifying the non-longitudinal propagation in the pink cluster; (e) \textit{SF1\_BlobSize\_Max} highlighting massive edema pooling in the turquoise cluster; and (f) \textit{GF\_Elongation} defining the streak-like infiltrates of the light blue cluster. The full set of 63 feature gradient maps is available in Supplementary Figure S1.}
		\label{fig:umap_parc}
	\end{figure*}

	Subsequent to the diagnostic projection, we aimed to objectively identify inherent structural subpopulations. Phenotyping by Accelerated Refined Classification (PARC) was applied to the manifold, revealing five distinct topological clusters (Figure \ref{fig:umap_parc}a). Furthermore, to decode the biological drivers of this separation, the values of individual spatial biomarkers were projected back onto the UMAP embeddings as gradient maps (Figure \ref{fig:umap_parc}b-f). This feature gradient mapping successfully isolates the specific architectural signatures—such as orientation, skeletal branching, and fragmentation—that define each disease sub-cohort. The complete repository of gradient maps for all 63 extracted features is provided in Supplementary Figure S1.
	
	\section{Discussion}
	The integration of morphological graph-radiomics advances muscle MRI evaluation from purely volumetric fat estimation into precise architectural profiling. While absolute fat quantity remains a critical baseline metric to assess general disease burden, relying solely on it obscures fundamental pathological differences between NMDs. Our 3D global methodology overcomes the limitations of classical radiomics and "black-box" models, providing an interpretable foundation for deep phenotyping.
	
	\subsection{Clinical Phenotype Translation and Biological Basis}
	The statistical findings and pairwise effect sizes derived from the heatmaps structurally validate observed clinical and biological realities. The pronounced effect sizes generated by global fragmentation metrics (\textit{SF2\_BlobSize\_Median}) effectively isolate specific phenotypes characterized by patchy, multi-focal infiltrates. This structurally mirrors the early clinical phase of the disease, frequently detected as muscle edema prior to irreversible fat replacement. 
	
	Conversely, topological markers like \textit{GF\_Skel\_Nodes} structurally validate whether the degeneration progresses via a uniform, cohesive, and continuous tissue replacement or a highly branched network. The visual translation of this phenomenon can be directly observed in our topological representations (Figure \ref{fig:graphs}), and is visually demonstrated by the UMAP projection, where DUX4 patients form a highly isolated cluster characterized by an extreme density of branching nodes. This distinct topological footprint objectively maps the varying degrees of inter-fascicular permeation across the whole muscle volume. It is important to note that while DUX4 variants are clinically characterized by severe left-right asymmetry, our bilateral averaging methodology did not mask this specific genotypic signature. Although averaging mathematically dilutes unilateral severity, the topological features—such as extreme skeletal branching density (\textit{Skel\_Nodes}) and distinct multi-focal fragmentation—are so morphologically pronounced in the affected limb that the patient's aggregated topological mean remains starkly distinct from symmetric pathologies. This proves that 3D morphological parameters capture the underlying nature of the architectural disruption, persisting powerfully even when fused into a single unified patient profile.
	
	Crucially, the engineered global biomarkers successfully separated phenotypes that are frequently challenging to distinguish using standard volumetry. For instance, the interface dynamics metrics (\textit{SF2\_To\_SF1\_Dist\_Min}) and center of mass parameters (\textit{SF1\_CoM\_Z}) generated notable effect sizes, indicating that spatial positioning and tissue proximity are defining characteristics of certain dystrophies. Since many variants belong to the Limb-Girdle Muscular Dystrophy (LGMD) spectrum, they frequently exhibit macroscopic similarities that confound standard visual grading. Variations in median blob sizes (\textit{GF\_BlobSize\_Median}) and geometric shapes effectively segregated these broader cohorts, demonstrating that extracting explicit 3D geometric signatures provides a robust mechanism to differentiate allied genetic variants.
	
	Furthermore, spatial vectoring parameters such as \textit{SF1\_CoM\_Z} effectively map the directional propagation of the disease. High differentiation power in this domain reflects whether the fat infiltration tracks longitudinally along the myofibers or distributes uniformly, highlighting the necessity of full 3D coordinate mapping over single-slice evaluations.
	
	\subsection{Dimensionality, Morphological Continuum, and Feature Gradients}
	The UMAP projection (Figure \ref{fig:umap}) based on the discriminative global features ($\epsilon^2 \ge 0.01$) reveals a distinct biological pattern regarding both phenotypic divergence and disease progression. Because UMAP operates as an unsupervised dimensionality reduction technique, it groups patients based purely on the mathematical similarity of their spatial biomarkers without prior knowledge of their clinical diagnoses. This intrinsic capability to cluster genetic cohorts highlights the robust diagnostic value of the engineered features. 
	
	For instance, the inclusion of biomarkers with smaller global effect sizes captured subtle structural variations. This allowed the algorithm to distinctly isolate the extreme architectural disruption of the DUX4 cohort into a separate topological "island," while also segregating a distinct sub-cluster of GNE cases. Meanwhile, the broad LGMD spectrum (CAPN3 and DYSF), alongside the remaining GNE variants, forms a more continuous main cluster. This overlap accurately reflects their shared overarching macroscopic similarities, typical for muscular dystrophies involving proximal limb segments, demonstrating that the topological profiling mirrors established clinical classifications.
	
	To characterize the specific 3D architectural signatures driving this phenotypic separation, we mapped individual spatial biomarkers onto the manifold using unsupervised PARC clustering (Figure \ref{fig:umap_parc}). Each of the five resulting clusters is defined by a unique morphological domain.
	
	The isolated dark blue cluster demonstrates the distinct phenotypic divergence of the DUX4 cohort, which is structurally validated by the direct feature mapping (Figure \ref{fig:dux4_diff}). Even as other diseases exhibit distinct morphological overlap across a continuous spectrum, unsupervised clustering consistently isolates DUX4 into an independent topological island. This spatial separation is dominated by extreme values of topological complexity (\textit{GF\_Skel\_Nodes}), reflecting a multi-focal network where the pathology invades the fascicles with dense, multidirectional branching. Conversely, the continuous LGMD cluster (including DYSF) exhibits a substantially lower node density, reflecting a more ordered, longitudinal infiltration pattern. Crucially, the validity of this mathematical isolation is corroborated by our pairwise post-hoc evaluations (Supplementary Table S2). Engineered features capturing network architecture generated exceptionally large effect sizes (frequently $r > 0.85$) strictly when differentiating DUX4 from all other genetic variants, supporting this highly branched 3D disruption as a key diagnostic marker of the disease.
	
	Conversely, the violet cluster is structurally distinguished by shifts in spatial vectoring, notably the center of mass (\textit{SF1\_CoM\_Z}, Figure \ref{fig:umap_parc}c). This indicates that early moderate fat replacement is localized significantly higher (proximal) or lower (distal) within the anatomical compartment, capturing a clear proximo-distal gradient of degeneration.
	
	The pink cluster is defined by atypically low geometric orientation values (\textit{GF\_MainAxis\_Z}, Figure \ref{fig:umap_parc}d). The infiltration in this sub-cohort lacks the typical longitudinal cylindrical growth; instead, the lipodegeneration propagates horizontally across inter-muscular fascial planes, resulting in a geometrically distinct, shortened vector of disease development.
	
	Furthermore, macroscopic fragmentation defines the turquoise cluster. Characterized by extreme values of fragment size (\textit{SF1\_BlobSize\_Max}, Figure \ref{fig:umap_parc}e), this group highlights patients where early fat and edema do not present as scattered spots. Instead, the pathology pools into massive, continuous, and merging macroscopic lakes prior to irreversible severe fat replacement.
	
	Finally, the broad morphological continuum of the light blue cluster is defined by high geometric elongation (\textit{GF\_Elongation}, Figure \ref{fig:umap_parc}f). Here, the fat explicitly propagates in long, narrow streaks, directly invading and replacing the longitudinal architecture of the surviving muscle fibers.
	
	As the pathology progresses, the unique geometric and topological signatures of each specific variant become morphologically dominant, driving the cohorts to diverge into distinct spatial regions. This proves that our continuous topological features capture fluid, highly specific disease states rather than static, spatially-agnostic volumetric categories.
	
	\subsection{Longitudinal Dynamics and Disease Progression}
	A critical advantage of extracting continuous, objective topological networks is their considerable potential in longitudinal tracking. Quantitative volumetry (such as Fat Fraction) provides only a static snapshot of the tissue ratio. In contrast, topological variables, such as the exponential growth rate of closed loops (\textit{GF\_Skel\_Cycles}) or the narrowing of the interface between fat replacement stages (\textit{SF2\_To\_SF1\_Dist\_Min}), allow clinicians to measure the spatial velocity and dynamics of the disease over time. Rather than utilizing these features exclusively for static cross-sectional differentiation, monitoring how the pathological "web" structurally permeates the fascicles (as visualized in Figure \ref{fig:graphs}c) can serve as a highly sensitive quantitative endpoint. This framework could evaluate the efficacy of emerging gene therapies by capturing active disease velocity and structural remodeling long before measurable overall muscle volume is lost.
	
	\subsection{Algorithmic Robustness and Interpretability}
	A notable limitation of contemporary machine learning frameworks (e.g., deep neural networks or ensemble methods like XGBoost) in rare disease diagnostics is their substantial data requirements and susceptibility to class imbalance. Predictive models frequently misclassify ultra-rare variants by favoring majority classes due to skewed training distributions. In contrast, by explicitly extracting topological invariants, our framework embeds 3D geometry directly into the mathematical formulation. Consequently, these metrics represent absolute physical properties (e.g., exact spatial distances in millimeters, specific branching node counts), making them less susceptible to cohort size discrepancies and robust against class imbalances.
	
	While advanced predictive systems rely on post-hoc explainability frameworks, such as SHAP values, to approximate the reasoning behind a "black-box" decision, our spatial vectoring methodology is intrinsically interpretable. Metrics such as the center of mass directly output the exact mathematical coordinates of pathological shifts without requiring secondary approximation layers, offering a robust and transparent foundation for clinical decision-making.
	
	\subsection{Limitations of the Proposed Methodology}
	Despite the considerable potential of morphological graph-radiomics, several limitations must be acknowledged. First, the extraction of 3D topological networks is inherently dependent on the quality of the initial tissue segmentation; inaccuracies in the severe or moderate fat replacement masks directly propagate to the skeletal graphs. Second, while the methodology models physical distances, highly anisotropic MRI acquisitions (where slice thickness significantly exceeds in-plane resolution) may distort the 3D skeletonization process, potentially introducing artifacts into the node and cycle counts. Moreover, while our topological invariants are theoretically hardware-agnostic, this study lacks external validation. Future studies must rigorously confirm the stability of these graph features across independent, multi-center cohorts involving diverse MRI vendors and magnetic field strengths (e.g., 1.5T vs. 3.0T).
	
	Furthermore, an important clinical consideration is the heterogeneity of the patient cohort regarding disease duration and clinical severity. Comprehensive clinical progression data (e.g., exact time since symptom onset) were not universally available, meaning our statistical evaluations are not strictly adjusted for disease stage. Consequently, the observed structural variations encapsulate a mixture of fundamental genotypic signatures and varying degrees of temporal disease progression. Finally, the computational complexity of computing exact Euclidean distance transforms and k-D tree-based graph networks across large 3D volumes is inherently higher than that of standard volumetric assessments.
	
	\section{Conclusion}
	This study presents a novel, topology-driven approach to muscle MRI analysis in neuromuscular disorders. By explicitly modeling the physical architecture of lipodegeneration (including structural network complexity, fragmentation, and interface dynamics), this methodology addresses the "bag-of-pixels" limitation of classical radiomics. By capturing the spatial heterogeneity that global Fat Fraction (FF) dilutes, these topological descriptors serve as a highly complementary extension to standard clinical volumetry. 
	
	The rigorous statistical evaluation confirmed that these global 3D descriptors yield substantial effect sizes. Furthermore, post-hoc analyses demonstrated that these spatial markers act as highly discriminative biomarkers, capable of isolating specific genetic phenotypes based on their unique macroscopic infiltration patterns. Ultimately, this morphological approach bridges the interpretability gap associated with black-box deep learning models, providing clinicians with a transparent, robust, and geometrically accurate foundation for deep radiomic phenotyping and differential diagnosis.
	
	Future investigations must address the realities of clinical deployment. To overcome the bottleneck of manual 3D segmentation, the integration of semi-supervised learning could leverage vast amounts of unannotated, raw scans. Moreover, to create a holistic Clinical Decision Support System (CDSS), these topological features should undergo multi-modal data fusion, integrating imaging biomarkers with clinical metadata (e.g., age, sex, genetic profiles) and extending the analysis to upper-body musculature to prevent accuracy degradation as new phenotypes are introduced. Transitioning to real-world hospital environments also necessitates robust strategies for missing data imputation, as comprehensive multi-sequence MRI protocols are frequently incomplete. Finally, to accumulate sufficient statistical power for rare diseases without compromising patient privacy, the establishment of federated learning networks across international consortiums will be paramount for the next generation of rare disease diagnostics.
	
	\section*{Code Availability}
	The automated feature extraction pipeline, including the 3D graph-based topology modules, was developed as an integral component of the MUSCAT (MUSCle fAt Topology) library. The codebase and the associated feature engineering algorithms are available from the corresponding author upon reasonable request.
	
	\section*{Acknowledgments}
	This research was supported by the CoMPaSS-NMD project (Computational Models for new Patients Stratification Strategies of Neuromuscular Disorders), funded by the HORIZON RIA program (HORIZON-HLTH-2022-TOOL-12-two-stage) under Grant Agreement No. GAP-101080874. We gratefully acknowledge their invaluable support and data provision.
	
	\bibliographystyle{unsrt}
	\bibliography{references}
	
\end{document}